\documentclass[lettersize,journal]{IEEEtran} 
\usepackage{amsmath,amssymb,amsfonts}
\usepackage{algorithmic}
\usepackage{algorithm}
\usepackage{array}
\usepackage[caption=false,font=normalsize,labelfont=sf,textfont=sf]{subfig}
\usepackage{textcomp}
\usepackage{stfloats}
\usepackage{url}
\usepackage{verbatim}
\usepackage{graphicx}
\usepackage{cite}
\usepackage{amssymb}  
\usepackage{amsmath}  
\usepackage[english]{babel}
\usepackage{amsmath}
\graphicspath{{./figures/}}   
\usepackage[rightcaption]{sidecap}
\usepackage{wrapfig}  
\usepackage{booktabs}    
\usepackage[colorlinks=true, allcolors=blue]{hyperref}
\usepackage[T1]{fontenc} 
\usepackage{multicol}    
\usepackage{multirow}   
\usepackage{longtable}
\usepackage{tabularx}
\usepackage{multirow}
\usepackage{booktabs}  
\usepackage{lscape}
\usepackage{array} 
\usepackage{makecell} 
\usepackage{longtable}
\usepackage[normalem]{ulem}
\usepackage{xurl}    
\usepackage{color}
\usepackage{xcolor}
\usepackage{lipsum}

\begin{document}
\title{Collaborative Perception Datasets for Autonomous Driving: A Review}

\author{Naibang~Wang, Deyong~Shang, Yan~Gong, Xiaoxi~Hu, Ziying~Song, Lei~Yang, Yuhan~Huang, Xiaoyu~Wang, Jianli~Lu \thanks{Naibang~Wang, Deyong~Shang, and Jianli~Lu are with the School of Mechanical and Electrical Engineering, China University of Mining and Technology (Beijing), Beijing, 100083, China. (email: sqt2300401025@student.cumtb.edu.cn; lujianli364@163.com).}\thanks{Yan~Gong is with the State Key Laboratory of Robotics and System, Harbin Institute of Technology, Harbin 150001, China. (email: gongyan2020@foxmail.com).}\thanks{Xiaoxi~Hu is with the State Key Laboratory of Intelligent Green Vehicle and Mobility, Tsinghua University, Beijing 100084, China. (email: xiaoxihurail@gmail.com).}\thanks{Ziying~Song is with Beijing Key Laboratory of Traffic Data Mining and Embodied Intelligence, School of Computer Science and Technology, Beijing Jiaotong University. (email: songziying@bjtu.edu.cn).}\thanks{Lei~Yang is with the School of Mechanical and Aerospace Engineering, Nanyang Technological University, Singapore. (email: yangleils@outlook.com).}\thanks{Yuhan~Huang is with the School of Mechatronics Engineering, Harbin Institute of Technology, 92 West Dazhi St., Harbin, China. (email: 23S108172@hit.edu.cn).}\thanks{Xiaoyu~Wang is with the Department of Electronic \& Electrical Engineering, University of Bath, UK. (email: xw2285@bath.ac.uk).}
}

\markboth{Journal of \LaTeX\ Class Files,~Vol.~14, No.~15, April~2025}%
\IEEEpubid{}

\maketitle
\begin{abstract}
Collaborative perception has attracted growing interest from academia and industry due to its potential to enhance perception accuracy, safety, and robustness in autonomous driving through multi-agent information fusion. With the advancement of Vehicle-to-Everything (V2X) communication, numerous collaborative perception datasets have emerged, varying in cooperation paradigms, sensor configurations, data sources, and application scenarios. However, the absence of systematic summarization and comparative analysis hinders effective resource utilization and standardization of model evaluation. As the first comprehensive review focused on collaborative perception datasets, this work reviews and compares existing resources from a multi-dimensional perspective. We categorize datasets based on cooperation paradigms, examine their data sources and scenarios, and analyze sensor modalities and supported tasks. A detailed comparative analysis is conducted across multiple dimensions. We also outline key challenges and future directions, including dataset scalability, diversity, domain adaptation, standardization, privacy, and the integration of large language models. To support ongoing research, we provide a continuously updated online repository of collaborative perception datasets and related literature: \url{https://github.com/frankwnb/Collaborative-Perception-Datasets-for-Autonomous-Driving}.
\end{abstract}

\begin{IEEEkeywords}
Autonomous Driving, Collaborative Perception, Datasets, V2X Communication, Perception Tasks
\end{IEEEkeywords}

\section{Introduction} \label{sec:introduction}
Perception is fundamental to Autonomous Driving (AD) systems~\cite{fan2023autonomous},~\cite{sun2020scalability},~\cite{huang2022multi}, \cite{li2020deep},~\cite{li2020lidar},~\cite{mao20233d},~\cite{velasco2020autonomous},~\cite{gong2023sifdrivenet}, enabling vehicles to understand their surroundings via onboard sensors for tasks such as object detection~\cite{wang2023sat},~\cite{song2023graphalign}, semantic segmentation, tracking~\cite{su2024cooperative}, and trajectory prediction~\cite{gong2023skipcrossnets}. The accuracy and robustness of perception directly affect safety and decision-making. Recent advances in computer vision~\cite{ren2015faster}, \cite{han2022survey}, \cite{khan2022transformers}, deep learning~\cite{grigorescu2020survey},~\cite{guo2020deep}, and sensor technology~\cite{bijelic2020seeing}, \cite{abdulmaksoud2025transformer},~\cite{yang2022binocular} have led to substantial progress in single-vehicle perception, leveraging cameras~\cite{wang2019pseudo}, \cite{li2024bevformer}, LiDAR~\cite{shi2019pointrcnn},~\cite{lang2019pointpillars}, radar~\cite{palffy2020cnn}, \cite{zheng2022tj4dradset}, and sensor fusion~\cite{song2022msfyolo},~\cite{li2022deepfusion},~\cite{song2025graphbev}, \cite{chen2024lvp}, \cite{wang2022interfusion}. However, these systems are constrained by limited field of view, sensor blind spots, and occlusions, making them insufficient in complex, dynamic traffic scenarios.

\begin{figure}[t]
\centerline{\includegraphics[width=1\linewidth]{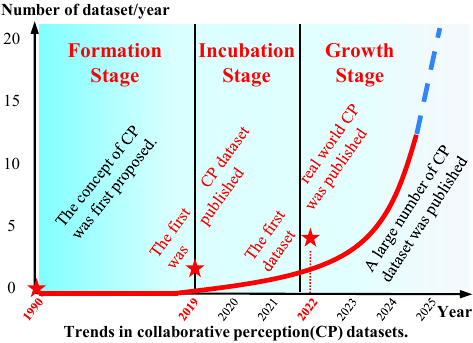}}
\caption{Development Trends of Collaborative Perception Datasets.}
\label{fig01}
\end{figure}

To address these limitations, Collaborative Perception (CP) leverages Vehicle-to-Everything (V2X) communication to enable real-time data sharing among vehicles and infrastructure~\cite{gao2024survey}, thereby enhancing the perception of occluded and long-range regions, improving safety, and expanding the Operational Design Domain~\cite{xu2022v2x}, \cite{hu2022where2comm}. For example, at urban intersections, a vehicle may fail to detect occluded agents; V2X links allow perception data exchange with infrastructure or other vehicles to overcome such blind spots. V2X includes Vehicle-to-Vehicle (V2V), Vehicle-to-Infrastructure (V2I), Infrastructure-to-Infrastructure (I2I), and Vehicle-to-Network modes, enabling more comprehensive environmental awareness.

High-quality datasets are critical for algorithm development and benchmarking in CP~\cite{yu2022dair}. Traditional datasets like KITTI~\cite{Geiger2012CVPR}, nuScenes~\cite{caesar2020nuscenes}, OpenMPD~\cite{zhang2022openmpd}, and Waymo~\cite{sun2020scalability} focus on single-agent perception and lack support for inter-agent communication, making them unsuitable for CP research. Recently, numerous CP datasets, as illustrated in Fig.~\ref{fig01}, have been released to support various scenarios, tasks, and sensor modalities, including DAIR-V2X~\cite{yu2022dair}, OPV2V~\cite{xu2022opv2v}, V2X-Sim~\cite{li2022v2x}, and V2X-Seq~\cite{yu2023v2x}.  These datasets vary in scale, annotation quality, sensor layout, and latency modeling. A systematic review is necessary to clarify their characteristics, limitations, and application scope, and to inform future research directions in collaborative perception.

\subsection{Roadmap}

\begin{figure*}[t]
\centerline{\includegraphics[width=1\linewidth]{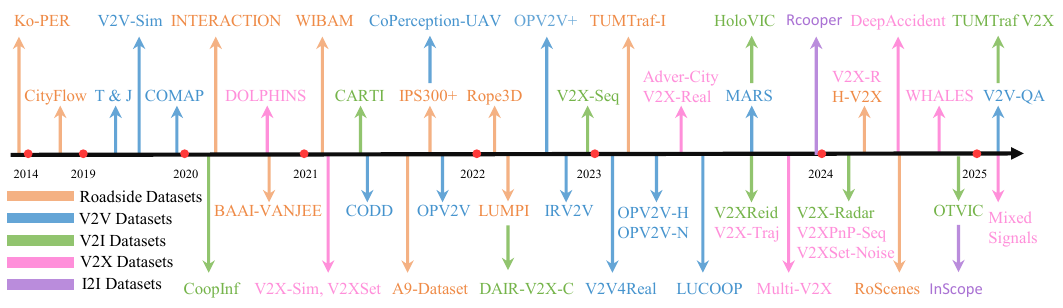}}
\caption{Chronological overview of collaborative perception datasets, including roadside, V2V, V2I, V2X, and I2I datasets.}
\label{fig1}
\end{figure*}

Fig.~\ref{fig1} summarizes the chronological evolution of CP datasets across major collaboration paradigms. Development has progressed from static roadside perception to dynamic multi-agent collaboration, driven by increasing task complexity, richer sensors, and advanced communication technologies. Initial datasets, such as Ko-PER~\cite{strigel2014ko}, CityFlow~\cite{tang2019cityflow}, and INTERACTION~\cite{zhan2019interaction}, focused on roadside perception without inter-agent communication. Around 2020, datasets like T\&J~\cite{chen2019cooper}, V2V-Sim~\cite{wang2020v2vnet}, CoopInf~\cite{arnold2020cooperative}, and CODD~\cite{arnold2021fast} introduced V2V and V2I collaboration, though often under constrained settings. Later efforts expanded scenario coverage: Adver-City~\cite{karvat2024adver} targets adverse weather, DeepAccident~\cite{wang2024deepaccident} supports motion and accident prediction, and H-V2X~\cite{liu2024h} covers highway scenes. 

Since 2022, the focus has shifted to real-world deployment and sensor diversity. Datasets such as V2V4Real~\cite{xu2023v2v4real}, V2X-Real~\cite{xiang2025v2x}, and DAIR-V2X-C~\cite{yu2022dair} emphasize realism, while V2X-Seq~\cite{yu2023v2x}, V2XReID~\cite{wang2024dair}, and DAIR-V2X-Traj~\cite{ruan2025learning} extend support to trajectory and identity tasks. V2X-Radar~\cite{yang2024v2x} and V2X-R~\cite{huang2024v2x} incorporate 4D radar, and V2XSet-Noise~\cite{meng2024agentalign} simulates deployment noise for robustness benchmarking. Recent I2I datasets such as InScope~\cite{zhang2024inscope} and RCooper~\cite{hao2024rcooper} further diversify collaboration modes. Overall, CP datasets are evolving toward higher realism, richer modalities, and more specialized tasks.

\subsection{Comparison to Related Reviews}

This review distinguishes itself from prior reviews~\cite{wang2024survey}, \cite{liu2024survey}, \cite{zhang2023multi}, \cite{song2023synthetic}, \cite{yazgan2024survey}, \cite{xiang2023multi}, \cite{malik2021collaborative}, \cite{liu2023towards}, \cite{huang2023v2x}, \cite{gao2024vehicle}, \cite{han2023collaborative}, which primarily focus on vehicle-centric datasets, specific techniques, or particular fusion strategies. While a few works partially overlap with ours in dataset coverage~\cite{teufel2024collective}, \cite{yazgan2024collaborative}, they overlook several recent CP datasets and benchmarks, and often lack systematic categorization or comparative depth. In contrast, our review provides a comprehensive and up-to-date compilation of CP datasets, with detailed analysis of their diversity, limitations, and suitability across various CP scenarios.

\begin{figure*}[t]
\centerline{\includegraphics[width=1\linewidth]{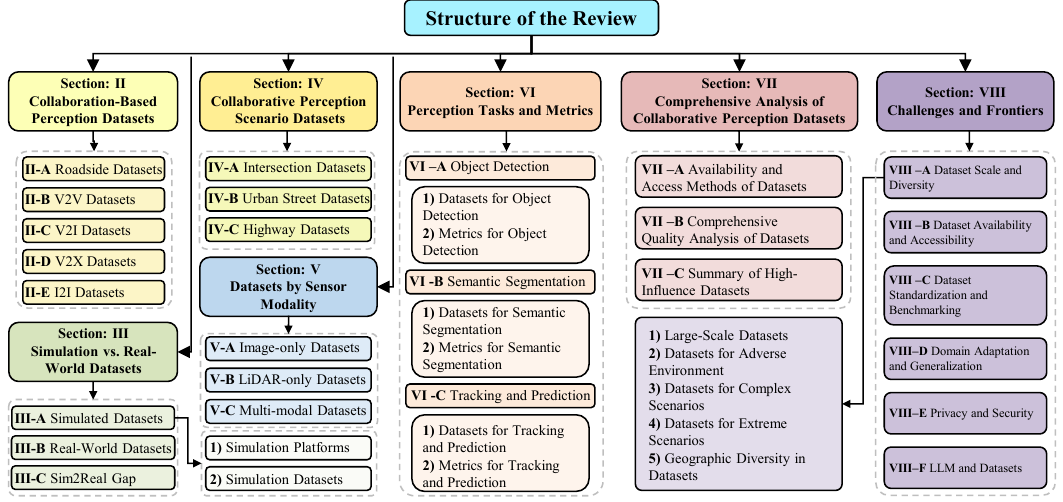}}
\caption{\textcolor{black}{Organization of this paper.}}
\label{fig2}
\end{figure*}

\subsection{Main Contributions}
The key contributions are summarized as follows:

\textbf{Comprehensive Dataset Review:} To the best of our knowledge, this is the first dedicated review that focuses exclusively on CP datasets. We systematically categorize and comprehensively analyze the existing datasets based on collaboration paradigms, data sources, sensor modalities, and application scenarios.

\textbf{Task-Centric Classification:} We organize datasets by core perception tasks in AD, including 3D object detection, semantic segmentation, and object tracking and prediction. This task-oriented taxonomy aids researchers in selecting appropriate datasets for specific applications. Recent developments and newly released datasets are also included to ensure timeliness.

\textbf{Comparative Analysis and Future Trends:} We conduct multi-dimensional comparisons, highlight representative datasets, and discuss emerging challenges such as standardization, LLM-based annotation, and cross-domain diversity, providing insights for future CP research.

\subsection{Review Structure}

The organization of this paper is illustrated in Fig.~\ref{fig2}. Section~\ref{sec:2} introduces CP datasets grouped by collaboration paradigms. Section~\ref{sec:3} compares simulated and real-world datasets. Section~\ref{sec:4} focuses on application scenarios, while Section~\ref{sec:5} analyzes sensor modalities. Section~\ref{sec:6} covers core perception tasks and evaluation metrics. Section~\ref{sec:7} provides dataset-level comparisons. Section~\ref{sec:8} discusses key challenges and trends. Section~\ref{sec:9} concludes the paper. A summary of abbreviations is provided in Table~\ref{tab1:Abbreviation}.

\begin{table}[htbp]
\centering
\caption{Summary of commonly used abbreviations in this review.}
\label{tab1:Abbreviation}
\begin{tabular}{ll}
\toprule
\textbf{Abbreviation} & \textbf{Definition} \\
\midrule
AD           & Autonomous Driving                      \\
AV           & Autonomous Vehicle                      \\
ICVs         & Intelligent Connected Vehicles \\
CAV          & Connected and Autonomous Vehicle        \\
CP           & Cooperative or Collaborative Perception \\
V2X          & Vehicle-to-Everything \\
V2V          & Vehicle-to-Vehicle \\
V2I          & Vehicle-to-Infrastructure \\
I2I          & Infrastructure-to-Infrastructure \\
ITS          & Intelligent Transportation System \\
BEV          & Bird’s Eye View \\
MF           & Motion Forecasting \\
Re-ID        & Re-Identification \\
LLM          & Large Language Model \\

\bottomrule
\end{tabular}
\end{table}

\section{Collaboration-Based Perception Datasets} \label{sec:2}

\begin{figure*}[t]
\centerline{\includegraphics[width=1\textwidth]{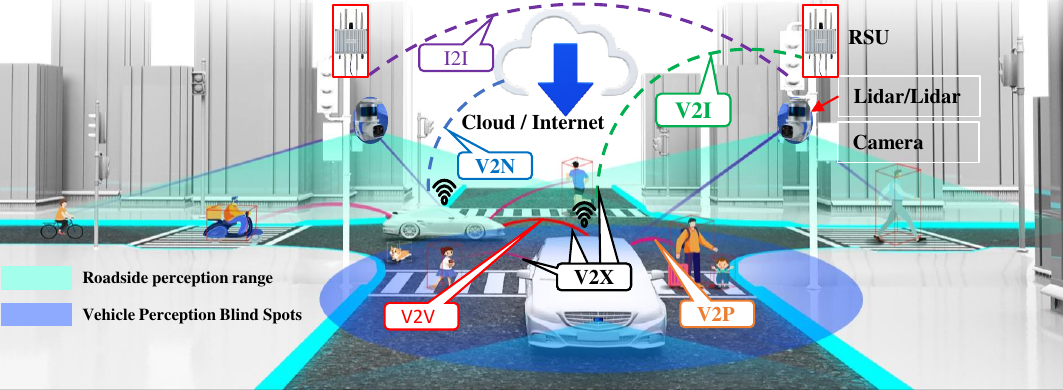}}
\caption{Schematic diagram of different collaborative perception methods, including roadside perception, V2V, V2I, V2X, and I2I.}
\label{fig3}
\end{figure*}

With the advancement of Intelligent Connected Vehicles (ICVs) and Intelligent Transportation Systems (ITS), CP paradigms have expanded beyond roadside perception to include V2V, V2I, V2X, and the emerging I2I frameworks. As illustrated in Fig.~\ref{fig3}, these paradigms support diverse communication and fusion architectures. This section categorizes CP datasets by collaboration type: roadside (Section~\ref{sec:2.1}), V2V (Section~\ref{sec:2.2}), V2I (Section~\ref{sec:2.3}), V2X (Section~\ref{sec:2.4}), and I2I (Section~\ref{sec:2.5}). Table~\ref{tab:dataset_comparison} summarizes key datasets in each category. These resources are critical to driving progress in cooperative intelligence for next-generation ICVs and ITS.

\begin{table*}[htbp]
\centering
\caption{Publicly Available Datasets for Collaborative Perception}
\label{tab:dataset_comparison}
\renewcommand{\arraystretch}{1.2} 
\small
\begin{tabularx}{\linewidth}{@{}p{0.6cm} l c c c c c c c c c@{}}
\toprule
\multirow{2}{*}{\textbf{View}} 
& \multirow{2}{*}{\textbf{Dataset}} 
& \multirow{2}{*}{\textbf{Year}} 
& \multirow{2}{*}{\textbf{Venue}} 
& \multirow{2}{*}{\textbf{Sensors}} 
& \multirow{2}{*}{\textbf{Source}} 
& \multirow{2}{*}{\textbf{Tasks}} 
& \multicolumn{3}{c}{\textbf{Size}} 
& \multirow{2}{*}{\textbf{Link}} \\

\cmidrule(lr){8-10}
& & & & & & & \textbf{Image} & \textbf{LiDAR} & \textbf{3D Box} & \\

\midrule
\multirow{10}{*}{Infra}
& Ko-PER~\cite{strigel2014ko} & 2014 & ITSC & C, L & Real & 3DOD, MOT & 18.7k & 4.8k & - & \href{https://www.uni-ulm.de/in/mrm/forschung/datensaetze.html}{Link} \\
& CityFlow~\cite{tang2019cityflow} & 2019 & CVPR & C & Real & MTSCT/MTMCT, ReID & 118k & - & 22.9k & \href{https://cityflow-project.github.io/}{Link} \\
& INTERACTION~\cite{zhan2019interaction} & 2019 & IROS & C, L & Real & 2DOD, TP & 1.4M & - & 1.4M & \href{https://interaction-dataset.com/}{Link} \\
& A9-Dataset~\cite{cress2022a9} & 2022 & IV & C, L & Real & 3DOD & 1.1k & - & 14.4k & \href{https://a9-dataset.com/}{Link} \\
& IPS300+~\cite{wang2022ips300+} & 2022 & ICRA & C, L & Real & 2DOD, 3DOD & 56.7k & 14.2k & 4.5M & \href{http://www.openmpd.com/column/IPS300}{Link} \\
& Rope3D~\cite{ye2022rope3d} & 2022 & CVPR & C, L & Real & 2DOD, 3DOD & 50k & - & 1.5M & \href{https://cloud.tsinghua.edu.cn/f/0cf0ed190d3e45428a0c/?dl=1}{Link} \\

& LUMPI~\cite{busch2022lumpi} & 2022 & IV & C, L & Real & 3DOD & 200k & 90k & - & \href{https://data.uni-hannover.de/cs_CZ/dataset/lumpi}{Link} \\
& TUMTraf-I~\cite{zimmer2023tumtraf} & 2023 & ITSC & C, L & Real & 3DOD & 4.8k & 4.8k & 57.4k & \href{https://innovation-mobility.com/en/project-providentia/a9-dataset/}{Link} \\
& RoScenes~\cite{zhu2024roscenes} & 2024 & ECCV & C & Real & 3DOD & 1.3M & - & 21.13M & \href{https://roscenes.github.io./}{Link} \\
& H-V2X~\cite{liu2024h} & 2024 & ECCV & C, R & Real & BEV Det, MOT, TP & 1.94M & - & - & \href{https://pan.quark.cn/s/86d19da10d18}{Link} \\
\midrule
\multirow{10}{*}{V2V}
& COMAP~\cite{yuan2021comap} & 2021 & ISPRS & L, C & Sim & 3DOD, SS & 8.6k & - & 226.9k & \href{https://demuc.de/colmap/}{Link} \\
& CODD~\cite{arnold2021fast} & 2021 & RA-L & L & Sim & Registration & - & 13.5k & 204k & \href{https://github.com/eduardohenriquearnold/fastreg}{Link} \\
& OPV2V~\cite{xu2022opv2v} & 2022 & ICRA & C, L, R & Sim & 3DOD & 44k & 11.4k & 232.9k & \href{https://mobility-lab.seas.ucla.edu/opv2v/}{Link} \\
& OPV2V+~\cite{hu2023collaboration} & 2023 & CVPR & C, L, R & Sim & 3DOD & 11.4k+ & - & 232.9k+ & \href{https://siheng-chen.github.io/dataset/CoPerception+/}{Link} \\
& V2V4Real~\cite{xu2023v2v4real} & 2023 & CVPR & L, C & Real & 3DOD, MOT, S2R & 40k & 20k & 240k & \href{https://mobility-lab.seas.ucla.edu/v2v4real/}{Link} \\
& LUCOOP~\cite{axmann2023lucoop} & 2023 & IV & L & Real & 3DOD & - & 54k & 7k & \href{https://data.uni-hannover.de/vault/icsens/axmann/lucoop-leibniz-university-cooperative-perception-and-urban-navigation-dataset/}{Link} \\
& MARS~\cite{li2024multiagent} & 2024 & CVPR & L, C & Real & VPR, NR & 1.4M & - & - & \href{https://ai4ce.github.io/MARS/}{Link} \\
& OPV2V-H~\cite{lu2024extensible} & 2024 & ICLR & C, L, R & Sim & 3DOD & 79k & - & 232.9k+ & \href{https://github.com/yifanlu0227/HEAL}{Link} \\
& V2V-QA~\cite{chiu2025v2v} & 2025 & arXiv & L, C & Real & 3DOD, PQA & - & 18k & - & \href{https://eddyhkchiu.github.io/v2vllm.github.io/}{Link} \\
& CP-UAV~\cite{hu2022where2comm} & 2022 & NIPS & L, C & Sim & 3DOD & 131.9K & - & 1.94M & \href{https://siheng-chen.github.io/dataset/coperception-uav/}{Link} \\

\midrule
\multirow{8}{*}{V2I}
& CoopInf~\cite{arnold2020cooperative} & 2020 & TITS & L, C & Sim & 3DOD & 10k & - & 121.2K & \href{https://github.com/eduardohenriquearnold/coop-3dod-infra?tab=readme-ov-file}{Link} \\
& DAIR-V2X-C~\cite{yu2022dair} & 2022 & CVPR & L, C & Real & 3DOD & 71k & 71k & 1.2M & \href{https://air.tsinghua.edu.cn/DAIR-V2X/index.html}{Link} \\
& V2X-Seq~\cite{yu2023v2x} & 2023 & CVPR & L, C & Real & 3DOT, TP & 71k & 15k & 464k & \href{https://github.com/AIR-THU/DAIR-V2X-Seq}{Link} \\
& HoloVIC~\cite{ma2024holovic} & 2024 & CVPR & L, C & Real & 3DOD, MOT & 100k & - & 11.47M & \href{https://holovic.net}{Link} \\
& OTVIC~\cite{zhu2024otvic} & 2024 & IROS & L, C & Real & 3DOD & 15k & - & 24.4k & \href{https://sites.google.com/view/otvic}{Link} \\
& DAIR-V2XReid~\cite{wang2024dair} & 2024 & TITS & L, C & Real & 3DOD, Re-ID & 7.3k & - & - & \href{https://github.com/Niuyaqing/DAIR-V2XReid}{Link} \\
& TUMTraf V2X~\cite{zimmer2024tumtraf} & 2024 & CVPR & L, C & Real & 3DOD, MOT & 5k & 2k & 30k & \href{https://tum-traffic-dataset.github.io/tumtraf-v2x/}{Link} \\
& V2X-Radar~\cite{yang2024v2x} & 2024 & arxiv & L, C, R & Real & 3DOD & 40k~ & 20k & 350k & \href{http://openmpd.com/column/V2X-Radar}{Link} \\

\midrule
\multirow{13}{*}{V2X}
& V2X-Sim~\cite{li2022v2x} & 2022 & RA-L & L, C & Sim & 3DOD, MOT, SS & 60k & 10K & 26.6K & \href{https://ai4ce.github.io/V2X-Sim/download.html}{Link} \\
& V2XSet~\cite{xu2022v2x} & 2022 & ECCV & L, C & Sim & 3DOD & 44k & 11.4k & 233k & \href{https://paperswithcode.com/dataset/v2xset}{Link} \\
& DOLPHINS~\cite{mao2022dolphins} & 2022 & ACCV & L, C & Sim & 2DOD, 3DOD & 42.3k & 42.3k & 292.5k & \href{https://dolphins-dataset.net/}{Link} \\
& V2XSet-Noise~\cite{meng2024agentalign} & 2024 & arXiv & L, C & Sim & 3DOD & 11.4k & 11.4k & - & \href{https://arxiv.org/abs/2412.06142}{Link} \\
& DeepAccident~\cite{wang2024deepaccident} & 2024 & AAAI & L, C & Sim & 3DOD, MOT, SS, TP & - & 57k & 285k & \href{https://deepaccident.github.io/}{Link} \\
& V2X-Real~\cite{xiang2025v2x} & 2024 & ECCV & L, C & Real & 3DOD & 171k & 33k & 1.2M & \href{https://mobility-lab.seas.ucla.edu/v2x-real}{Link} \\
& Multi-V2X~\cite{li2024multi} & 2024 & arxiv & L, C & Sim & 3DOD, MOT & 549k & 146k & 4.2M & \href{http://github.com/RadetzkyLi/Multi-V2X}{Link} \\
& Adver-City~\cite{karvat2024adver} & 2024 & arxiv & L, C & Sim & 3DOD, MOT, SS & 24k & 24k & 890k & \href{https://labs.cs.queensu.ca/quarrg/datasets/adver-city/}{Link} \\
& DAIR-V2X-Traj~\cite{ruan2025learning} & 2024 & NIPS & L, C & Real & MF & 808k & - & 1.4M & \href{https://github.com/AIR-THU/V2X-Graph}{Link} \\
& WHALES~\cite{chen2024whales} & 2024 & arxiv & L, C & Sim & 3DOD & 70k & 17k & 2.01M & \href{https://github.com/chensiweiTHU/WHALES}{Link} \\
& V2X-R~\cite{huang2024v2x} & 2024 & arxiv & L, C, R & Sim & 3DOD & 150.9k & 37.7k & 170.8k & \href{https://github.com/ylwhxht/V2X-R}{Link} \\
& V2XPnP-Seq~\cite{zhou2024v2xpnp} & 2024 & arxiv & L, C & Real & Perception and Prediction & 208k & 40k & 1.45M & \href{https://mobility-lab.seas.ucla.edu/v2xpnp/}{Link} \\
& SCOPE~\cite{gamerdinger2024scope} & 2024 & arxiv & C, L & Sim & 2DOD, 3DOD, SS, S2R & 17k & 17k & 575k & \href{https://ekut-es.github.io/scope}{Link} \\
& Mixed Signals~\cite{luo2025mixed} & 2025 & arxiv & L & Real & 3DOD & - & 45.1k & 240.6k & \href{https://mixedsignalsdataset.cs.cornell.edu/}{Link} \\
& Griffin~\cite{wang2025griffin} & 2025 & arXiv & C, L & Sim & 3DOD, MOT & 275k & - & - & \href{https://github.com/wang-jh18-SVM/Griffin}{Link} \\

\midrule
\multirow{2}{*}{I2I}
& Rcooper~\cite{hao2024rcooper} & 2024 & CVPR & C, L & Real & 3DOD, MOT & 50k & 30k & 30k & \href{https://github.com/AIR-THU/DAIR-Rcooper}{Link} \\
& InScope~\cite{zhang2024inscope} & 2024 & arxiv & L & Real & 3DOD, MOT & - & 21.3k & 188k & \href{https://github.com/xf-zh/InScope}{Link} \\

\bottomrule
\end{tabularx}

\vspace{1mm}
\begin{minipage}{\linewidth}
\footnotesize
\textbf{Note:} Sensors: Camera (C), LiDAR (L), Radar (R). Source: Real = collected in the real world; Sim = generated via simulation. Tasks: 2DOD = 2D Object Detection, 3DOD = 3D Object Detection, MOT = Multi-Object Tracking, MTSCT = Multi-target Single-camera Tracking, MTMCT = Multi-target Multi-camera Tracking, SS = Semantic Segmentation, TP = Trajectory Prediction, VPR = Visual Place Recognition, NR = Neural Reconstruction, Re-ID = Re-Identification, S2R = Sim2Real, MF = Motion Forecasting, PQA = Planning Q\&A.

\end{minipage}
\end{table*}

\subsection{Roadside Datasets}\label{sec:2.1}

Roadside perception plays a vital role in AD and ITS by offering a global, occlusion-resilient view of traffic participants~\cite{yang2024sgv3d},~\cite{zhao2024roadbev},~\cite{yang2024monogae}. Unlike vehicle-mounted systems, infrastructure-based perception units—typically installed on poles, gantries, or overpasses—enable the monitoring of complex intersections, dense traffic flows, and long-range environments with enhanced spatial stability and temporal continuity. By supplementing on-vehicle perception with infrastructure-level observations, roadside perception significantly improves perception robustness in scenarios involving occlusions, blind spots, or congested conditions. 

\begin{figure}[t]
\centering
\includegraphics[width=\linewidth]{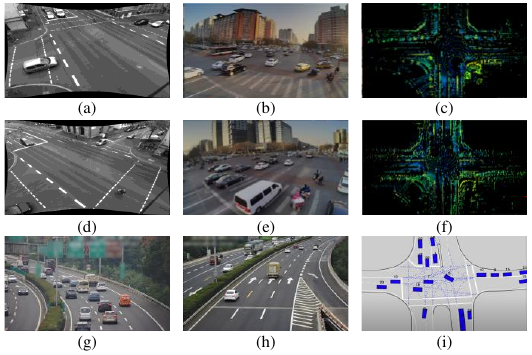}
\caption{Samples of roadside perception datasets. (a), (d) Ko-PER~\cite{strigel2014ko}; (b), (c), (e), (f) IPS300+~\cite{wang2022ips300+}; (g), (h) Rope3D~\cite{ye2022rope3d}; (i) INTERACTION~\cite{zhan2019interaction}.}
\label{fig4}
\end{figure}

A variety of representative roadside datasets have been introduced to support infrastructure-based perception research. Early datasets such as Ko-PER~\cite{strigel2014ko} and CityFlow~\cite{tang2019cityflow} primarily target intersections and Re-ID tasks. Later datasets expand coverage and functionality: INTERACTION~\cite{zhan2019interaction} emphasizes interactive behaviors using drone-captured trajectories and semantic maps; Rope3D~\cite{ye2022rope3d} provides top-down monocular images with 3D annotations; LUMPI~\cite{busch2022lumpi} and IPS300+~\cite{wang2022ips300+} support multi-view, multi-modal fusion with rich annotations; and TUMTraf-I~\cite{zimmer2023tumtraf}, A9-Dataset~\cite{cress2022a9}, RoScenes~\cite{zhu2024roscenes}, and H-V2X~\cite{liu2024h} extend coverage to diverse weather, lighting, and highway scenarios.

Fig.~\ref{fig4} illustrates the diversity across Ko-PER~\cite{strigel2014ko}, IPS300+~\cite{wang2022ips300+}, Rope3D~\cite{ye2022rope3d}, and INTERACTION~\cite{zhan2019interaction} in sensor modalities, viewpoints, and scene complexities. Ko-PER and Rope3D use monocular or stereo RGB with fixed viewpoints; Rope3D emphasizes long-range top-down detection, while Ko-PER captures lateral mid-range views. IPS300+ integrates LiDAR and stereo cameras for dense, occlusion-resilient 3D annotations. INTERACTION offers behavior-rich drone views for planning and imitation learning tasks. Together, they reflect trade-offs between perception complexity, spatial coverage, and task specialization.

These datasets have expanded infrastructure-centric perception research across detection, segmentation, and tracking tasks, while also enhancing V2I systems through broader scene coverage and perceptual redundancy. Nonetheless, challenges persist, including deployment costs, urban-centric biases, and difficulties in sensor synchronization and alignment across deployments.

\subsection{V2V Datasets}\label{sec:2.2}

V2V CP enhances perception coverage and accuracy by enabling vehicles to share information across viewpoints. It helps overcome occlusions and sensor limitations, supporting robust detection, semantic understanding, and motion prediction in multi-agent environments. High-quality datasets are thus vital for advancing V2V perception research.

V2V datasets have evolved significantly in alignment strategies and task diversity. Early datasets like OPV2V~\cite{xu2022opv2v}, LUCOOP~\cite{axmann2023lucoop}, and MARS~\cite{li2024multiagent} assume ideal synchronization via GPS or timestamps, overlooking practical challenges such as asynchrony and drift. Later datasets incorporate asynchronous modeling (HEAL~\cite{lu2024extensible}), neural alignment (OPV2V-N~\cite{wang2024rcdn}), and SLAM-based fusion (COMAP~\cite{yuan2021comap}). OPV2V+~\cite{hu2023collaboration} adds view selection and perception completion. While most support 2–4 vehicles, MARS emphasizes trajectory diversity. Tasks now cover detection, tracking, Bird’s Eye View (BEV) segmentation, and planning. V2V-QA~\cite{chiu2025v2v} further introduces multi-modal reasoning using LLMs; IRV2V~\cite{wei2024asynchrony} and CODD~\cite{arnold2021fast} focus on registration under asynchrony.

Fig.~\ref{fig5} (a–c) illustrates this progression. OPV2V (a) provides synthetic CARLA data for controlled and reproducible benchmarking but lacks real-world complexity. V2V4Real (b) contributes large-scale real LiDAR data involving dual agents in diverse scenarios. V2V-QA (c) extends this by introducing multi-modal reasoning tasks powered by LLMs. Collectively, these datasets mark a shift from basic sensor fusion toward semantically rich and cognitively informed V2V collaboration.

Despite these advances, most V2V datasets still rely on offline stitching and neglect realistic modeling of communication delay, bandwidth constraints, and uncertainty. Future efforts should integrate real-time communication, dynamic alignment, and large-scale multi-agent urban scenes to support deployable cooperative strategies.

\subsection{V2I Datasets}\label{sec:2.3}

V2I perception datasets enhance AD systems by facilitating collaboration between vehicles and roadside infrastructure. These datasets integrate multimodal information from both vehicle-mounted and infrastructure-based sensors, enabling a more accurate and holistic understanding of the driving environment.

Representative V2I datasets exhibit distinct characteristics and collaboration designs. CoopInf~\cite{arnold2020cooperative} utilizes eight infrastructure sensors to support multi-view 3D object detection. DAIR-V2X-Seq~\cite{yu2023v2x} is the first large-scale sequential V2X dataset incorporating time-aligned data from both vehicles and infrastructure. HoloVIC~\cite{ma2024holovic} creates holographic intersection scenes with diverse layouts, aiming to mitigate occlusion and blind spots through multimodal fusion. TUMTraf-V2X~\cite{zimmer2024tumtraf} focuses on perception in challenging lighting conditions and supports 3D detection and tracking tasks. OTVIC~\cite{zhu2024otvic} addresses high-speed and noisy communication settings using multimodal, multi-view sensor data.

\begin{figure}[t]
\centering
\includegraphics[width=\linewidth]{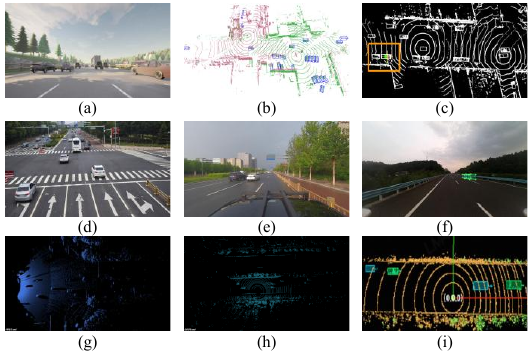}
\caption{Samples from representative V2V and V2I datasets. (a) OPV2V~\cite{xu2022opv2v}; (b) V2V4Real~\cite{xu2023v2v4real}; (c) V2V-QA~\cite{chiu2025v2v}; (d)(e)(g)(h) DAIR-V2X~\cite{yu2022dair}; (f)(i) OTVIC~\cite{zhu2024otvic}.}
\label{fig5}
\end{figure}

As illustrated in Fig.~\ref{fig5} (d)--(i), DAIR-V2X ((d)(e)(g)(h)) provides synchronized vehicle-side and infrastructure-side data across various lighting and weather conditions, making it suitable for robust V2I perception under real-world deployment constraints. In contrast, OTVIC ((f)(i)) focuses on sparse highway scenarios with long-range detection tasks, leveraging BEV projections and trajectory-level annotations. The visual samples highlight the contrast in scene layout, detection distance, and data density, reflecting complementary application targets.

While these datasets significantly contribute to improving V2I CP, challenges remain. Sensor synchronization is technically demanding, and many datasets offer limited scene diversity or weather coverage~\cite{zhu2025deep}. Future research should prioritize more heterogeneous scene types and realistic multi-view synchronization to enhance generalization across V2I deployments.

\subsection{V2X Datasets}\label{sec:2.4}

V2X-based CP enables rich environmental understanding by linking vehicles, infrastructure, and other agents through communication. This paradigm enhances perception completeness and safety in dynamic traffic environments.

Recent V2X datasets increasingly emphasize agent diversity and heterogeneous collaboration. Early datasets like V2X-SIM~\cite{li2022v2x} support basic CP tasks in structured settings. The DAIR-V2X series~\cite{yu2022dair, wang2024dair} extends to V2I collaboration with asymmetric sensor setups. Multi-agent dynamics are explored in V2XPnP~\cite{zhou2024v2xpnp} and DOLPHINS~\cite{mao2022dolphins}, which incorporate sequential reasoning and adverse weather. Adver-City~\cite{karvat2024adver}, DeepAccident~\cite{wang2024deepaccident}, and V2X-Real~\cite{xiang2025v2x} improve realism and environmental diversity, while V2XSet-Noise~\cite{meng2024agentalign} injects simulated sensor and timing noise to evaluate robustness under deployment-like conditions, and V2X-Radar~\cite{yang2024v2x} incorporates 4D radar for cross-modal fusion.

\begin{figure}[t]
\centering
\includegraphics[width=\linewidth]{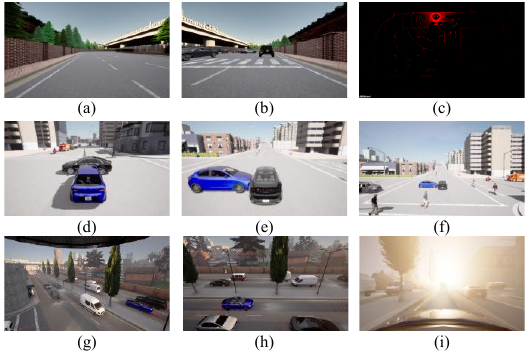}
\caption{Samples from representative V2X datasets. (a)(b)(c) V2VSet~\cite{xu2022v2x}; (d)(e)(f) DeepAccident~\cite{wang2024deepaccident}; (g)(h)(i) Adver-City~\cite{karvat2024adver}.}
\label{fig6}
\end{figure}

As shown in Fig.~\ref{fig6}, the three datasets reflect complementary directions in advancing V2X perception. V2VSet (a–c) offers clean synthetic environments for cooperative 3D detection and view fusion, but lacks dynamic risk modeling. DeepAccident (d–f) introduces accident-prone and motion-rich sequences to support forecasting tasks. Adver-City (g–i) emphasizes sensor degradation under fog, glare, and other adverse conditions. These samples illustrate a trade-off between controllability (V2VSet), temporal realism (DeepAccident), and visual robustness (Adver-City), jointly covering core challenges in V2X CP.

Despite progress, V2X datasets still face scalability and integration challenges. Many lack coverage of complex intersection layouts or hybrid traffic patterns. Benchmarks often isolate perception tasks, limiting support for end-to-end CP evaluation across heterogeneous agents and adverse conditions.

\subsection{I2I Datasets}\label{sec:2.5}

As V2X CP advances, I2I collaboration emerges as a key paradigm to address the limitations of single-agent roadside systems. While traditional V2I and V2X enhance perception via vehicle–infrastructure communication, they remain susceptible to occlusions from static objects or large vehicles. I2I mitigates this by linking multiple roadside perception units via low-latency channels, enabling real-time feature or decision-level fusion. This improves robustness in occlusion-prone scenes such as intersections and corridors, and supports scalable, decentralized cooperative frameworks.

Three recent datasets highlight advances in I2I perception. RCooper~\cite{hao2024rcooper} is the first large-scale real-world roadside dataset, covering intersection and corridor scenes with heterogeneous infrastructure agents using diverse LiDARs and cameras. It provides 3D detection and tracking labels to validate cross-node fusion. InScope~\cite{zhang2024inscope} emphasizes occlusion-aware perception using multi-view LiDARs and introduces a degradation metric for systematic evaluation. V2XPnP~\cite{zhou2024v2xpnp} complements these with long-sequence, multi-agent data—including I2I settings—to support temporal modeling and complex tasks.

\begin{figure}[t]
\centering
\includegraphics[width=\linewidth]{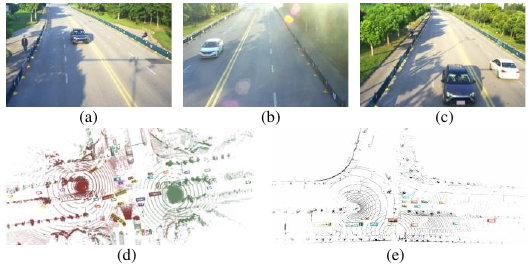}
\caption{Samples of I2I collaborative perception datasets. (a)--(d) RCooper~\cite{hao2024rcooper}; (e) InScope~\cite{zhang2024inscope}.}
\label{fig7}
\end{figure}

Fig.~\ref{fig7} compares two I2I datasets through representative samples. RCooper (a)--(d) features multi-angle RGB images and LiDAR point clouds from distributed infrastructure nodes, enabling long-range coverage and blind spot reduction via inter-node fusion. In contrast, InScope (e) employs a dual-LiDAR setup for occlusion-aware perception in dense urban scenes. While RCooper emphasizes spatial coverage and corridor-scale extensiveness, InScope focuses on precise detection in visually constrained environments, supported by its anti-occlusion evaluation metric. These datasets reflect complementary strategies in I2I perception: broad area fusion versus localized occlusion resolution.

Compared to other paradigms, I2I collaboration proves particularly effective in mitigating occlusions within dense traffic environments. However, current I2I datasets still face limitations in generalization, data synchronization, and scene diversity, underscoring the need for scalable deployment strategies and standardized evaluation protocols. Beyond I2I, emerging collaborative paradigms such as Vehicle-to-Network (V2N), cloud-based collaboration, and Vehicle-to-Pedestrian (V2P) are drawing increasing attention. V2N and cloud communication facilitate global map sharing and large-scale model updates, while V2P extends perception capabilities to vulnerable road users, broadening the scope and societal impact of CP systems.

\section{Simulation vs. Real-World Datasets} \label{sec:3}

While Section~\ref{sec:2} organized CP datasets according to communication paradigms such as V2V, V2I, and V2X, this section shifts perspective to the origin and fidelity of the data—specifically, the distinction between simulated and real-world datasets. This transition enables a deeper understanding of how dataset generation methods impact the development, evaluation, and deployment of CP systems.

Simulation and real-world datasets play complementary roles in collaborative perception. Simulated datasets offer scalable, controllable environments with dense annotations, enabling rapid iteration and benchmarking. In contrast, real-world datasets provide authentic sensor artifacts, diverse environmental conditions, and deployment realism essential for evaluating generalization and robustness. However, the domain gap between synthetic and real data introduces significant challenges for transferability and real-world performance~\cite{liu2024survey}. In this section, we first review mainstream simulation platforms and representative simulated datasets, then analyze widely used real-world datasets. We conclude by discussing the Sim2Real gap, a key challenge in CP research, where models trained in simulation often degrade when applied to real-world environments. Table~\ref{tab:sim_real_comparison} summarizes the core differences between simulated and real-world datasets, highlighting their respective advantages and limitations.

\begin{table}[htbp]
\centering
\caption{Comparison of Simulated and Real Datasets Characteristics}
\label{tab:sim_real_comparison}
\begin{tabular}{p{6.5cm} c c}

\toprule
\textbf{Characteristic} & \textbf{Sim} & \textbf{Real} \\

\midrule
Data Collection Cost & $\checkmark$ & $\times$ \\
Scene Diversity, Environmental Diversity & $\checkmark$ & $\times$ \\
Dynamic Interaction Complexity & $\times$ & $\checkmark$ \\
Ability to Simulate New Scenes & $\checkmark$ & $\times$ \\
Coverage of Extreme/Dangerous Scenes & $\checkmark$ & $\times$ \\
Coverage of Long-Tail Scenarios & $\checkmark$ & $\times$ \\
Difficulty in Large-Scale Data Collection & $\checkmark$ & $\times$ \\
Processing Complexity & $\times$ & $\checkmark$ \\
Data Labeling Accuracy & $\checkmark$ & $\times$ \\
Time Synchronization Precision & $\checkmark$ & $\times$ \\
Data Reality and Authenticity & $\times$ & $\checkmark$ \\
Data Randomness and Noise & $\times$ & $\checkmark$ \\
High Complexity and Diversity & $\times$ & $\checkmark$ \\
Repeatability, Algorithm Iteration Speed, High Safety & $\checkmark$ & $\times$ \\
High Cost of Updates and Maintenance & $\times$ & $\checkmark$ \\
Data Privacy Compliance & $\checkmark$ & $\times$ \\
Industry Certification and Recognition & $\times$ & $\checkmark$ \\

\bottomrule
\end{tabular}
\end{table}

\subsection{Simulated Datasets}\label{sec:3.1}
\subsubsection{Simulation Platforms}

Simulation platforms play a pivotal role in CP research for AD, providing scalable, controllable, and annotation-rich alternatives to costly real-world data collection. By enabling the generation of synthetic sensor data with fine-grained ground truth, these simulators accelerate the development and validation of perception algorithms under diverse and repeatable conditions.

Among open-source simulation platforms, CARLA~\cite{dosovitskiy2017carla} is one of the most widely adopted tools for AD research. It supports various multimodal sensors such as RGB cameras, LiDAR, radar, and depth maps, and provides configurable urban and rural scenarios. However, its LiDAR simulation has limitations in modeling realistic point cloud intensity and echo effects. SUMO~\cite{krajzewicz2012recent} focuses on microscopic traffic flow and is often used to simulate vehicle and pedestrian trajectories. When integrated with CARLA, it enhances multi-agent interaction realism. To facilitate CP and V2X studies, OpenCDA~\cite{xu2021opencda} integrates both platforms, offering standardized scenarios, communication protocols, and benchmarks for end-to-end testing. AirSim~\cite{shah2018airsim} is another widely used simulator for vision-centric tasks, providing photorealistic environments, though it offers limited support for LiDAR and complex traffic scenes.

\subsubsection{Simulation Datasets}

Simulation datasets play a fundamental role in CP research by enabling controlled data generation, automatic annotation, and repeatable evaluation under diverse or rare conditions. Most simulated datasets are built upon open-source platforms such as CARLA, SUMO, and LiDARsim, which support flexible scenario modeling and multi-sensor simulation.

Among them, COMAP~\cite{yuan2021comap} and OPV2V~\cite{xu2022opv2v} employ CARLA-SUMO co-simulation to create interactive, multi-agent urban driving environments. These datasets support key CP tasks, including 3D object detection, semantic segmentation, and trajectory prediction. The OPV2V series has been extended in OPV2V+~\cite{hu2023collaboration} and OPV2V-H~\cite{lu2024extensible}, introducing increased agent diversity and modeling asynchronous perception to reflect real-world temporal misalignment.

Several recent datasets focus on robustness under challenging environmental conditions. DOLPHINS~\cite{mao2022dolphins} and Adver-City~\cite{karvat2024adver} simulate adverse weather scenarios such as rain, fog, and snow, which are critical for evaluating sensor degradation. In parallel, V2X-Sim~\cite{li2022v2x}, Multi-V2X~\cite{li2024multi}, and DeepAccident~\cite{wang2024deepaccident} introduce large-scale cooperative settings with support for high-resolution sensor streams, complex agent coordination, and accident scenario modeling. Together, these datasets provide robust testbeds for benchmarking CP models, studying Sim2Real transfer, and analyzing communication-aware perception strategies.

\subsection{Real-World Datasets}

Real-world datasets are essential for validating CP systems under authentic sensor noise, traffic dynamics, and environmental complexity. DAIR-V2X~\cite{yu2022dair} is the first large-scale multimodal real-world dataset focusing on V2I cooperation, while V2X-Seq~\cite{yu2023v2x} extends it with long-sequence data and traffic signal information. V2V4Real~\cite{xu2023v2v4real} is the first large-scale real-world multimodal V2V dataset, offering valuable support for Sim2Real studies despite limited CAV diversity. LUCOOP~\cite{axmann2023lucoop}, collected with three CAVs using heterogeneous sensors, supports urban driving studies, while MARS~\cite{li2024multiagent} expands to four CAVs for improved stability and generalization. HoloVIC~\cite{ma2024holovic} and OTVIC~\cite{zhu2024otvic} focus on V2I setups, with OTVIC simulating real-time vehicle-RSU communication to reflect dynamic transmission challenges in infrastructure-assisted perception. These datasets collectively bridge the gap between simulated environments and field deployments, fostering more transferable and robust CP systems.

\subsection{Sim2Real Gap}
The Sim2Real gap refers to the performance degradation observed when models trained on synthetic data are deployed in real-world environments. This challenge is especially pronounced in CP, where heterogeneous sensors, complex traffic dynamics, and multi-agent coordination amplify domain discrepancies. Although simulation platforms offer controllable and annotation-rich environments, their sensor data often lacks realistic noise, occlusions, and temporal misalignment. Real-world deployments, in contrast, introduce GPS drift, communication latency, and inconsistent sensor configurations, resulting in both geometric and semantic mismatches that undermine model generalization.

To study this problem, several benchmark datasets have been developed. Simulated datasets such as OPV2V~\cite{xu2022opv2v} and V2XSet~\cite{xu2022v2x} enable multi-agent experiments but fall short in capturing real-world uncertainties. Real-world datasets like V2V4Real~\cite{xu2023v2v4real} and DAIR-V2X~\cite{yu2022dair} offer more realistic settings with diverse sensor setups. Empirical studies have quantified the Sim2Real gap across these datasets: S2R-ViT~\cite{li2024s2r} reports up to 40\% AP drop when transferring from OPV2V to V2V4Real under deployment noise; CoPEFT~\cite{wei2025copeft} improves AP@0.7 from 21.7\% to 41.8\% on DAIR-V2X via parameter-efficient tuning. DUSA~\cite{kong2023dusa} and CUDA-X~\cite{yin2025cuda} achieve 5–20\% AP gains through domain-alignment modules. V2X-ReaLO~\cite{V2X-ReaLO} further reveals a 13.2 mAP drop during online deployment, highlighting the impact of latency and pose noise.

Collectively, these findings show that the Sim2Real gap in CP stems from both data distribution shifts and system-level constraints. Addressing it requires more realistic simulation environments, domain adaptation techniques, and hybrid training protocols that bridge synthetic and real domains. Future research should also prioritize time-aware, uncertainty-aware modeling and develop standardized benchmarks for cross-domain evaluation to ensure robust real-world deployment.

\section{Cooperative perception Scenario Datasets}\label{sec:4}

\begin{figure}[htbp]
\centering
\includegraphics[width=0.9\linewidth]{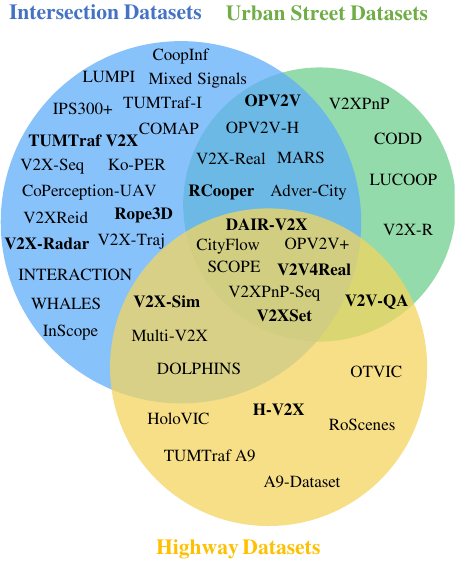}
\caption{A Venn diagram showing the distribution of CP datasets across intersection, urban street, and highway scenarios. The overlapping areas indicate datasets applicable to multiple scenarios.}
\label{fig8}
\end{figure}

Building on the discussion of data sources in Section~\ref{sec:3}, we now focus on the deployment contexts of CP datasets. Scenario-specific datasets are essential for evaluating system performance under diverse traffic conditions and supporting targeted algorithm development. This section categorizes CP datasets into three representative scenarios: intersections, urban streets, and highways. These settings differ in traffic complexity, agent density, and occlusion patterns, shaping dataset design and system requirements. Fig.~\ref{fig8} illustrates the distribution of representative datasets. For each category, we review key datasets, emphasizing their design motivations, sensor configurations, and task relevance.

\subsection{Intersection Datasets}
Intersections represent one of the most complex traffic environments, where vehicles, pedestrians, and cyclists converge from multiple directions, leading to significant challenges such as occlusions and dynamic interactions. To enhance perception capabilities and environmental understanding, intersection CP datasets typically employ multi-sensor fusion and cross-view information integration techniques. The following datasets are among the most representative in this category.

Ko-PER~\cite{strigel2014ko} is one of the earliest datasets using fixed LiDAR and cameras to observe intersection dynamics. IPS300+~\cite{wang2022ips300+} enhances resolution with dense 3D annotations, while LUMPI~\cite{busch2022lumpi} introduces multi-view data under diverse weather and lighting. TUMTraf-I~\cite{zimmer2023tumtraf} focuses on annotation precision in night and rain conditions. HoloVIC~\cite{ma2024holovic} designs holographic intersections to study sensor deployment. Several V2X datasets also incorporate intersection scenarios: DAIR-V2X~\cite{yu2022dair} supports V2I in urban crossings; DOLPHINS~\cite{mao2022dolphins} adds weather-varied intersections; DeepAccident~\cite{wang2024deepaccident} centers on accident prediction; V2XPnP~\cite{zhou2024v2xpnp} includes 163 intersections for trajectory prediction and tracking. These datasets emphasize high-precision perception in occlusion-prone intersections via multi-view perception and cross-agent fusion, though challenges like traffic variability and sensor blind spots remain.

\subsection{Urban Street Datasets}
Urban street CP datasets span V2V, V2I, and V2X paradigms, enabling tasks such as object detection, tracking, and trajectory prediction. CityFlow~\cite{tang2019cityflow} provides multi-camera tracking across arterial and residential roads. LUCOOP~\cite{axmann2023lucoop} uses synchronized multimodal data from three Connected and Autonomous Vehicles (CAVs), including 3D point clouds, city models, and UWB ranging. MARS~\cite{li2024multiagent} extends to four CAVs under diverse weather, enhancing collaborative robustness. Roadside-focused datasets also emerge: RCooper~\cite{hao2024rcooper} is the first large-scale real-world roadside CP dataset; RoScenes~\cite{zhu2024roscenes} offers dense multi-view 3D urban scenes for BEV perception; V2XPnP~\cite{zhou2024v2xpnp} captures multi-agent urban trajectories and interactions. Collectively, these datasets provide comprehensive urban traffic data for multi-agent perception, fusion, and prediction.

\subsection{Highway Datasets}

With the growing deployment of ICVs on highways, CP datasets for highway scenarios have become increasingly important for traffic management and AD safety. Highway environments demand long-range detection, lane-level perception, and robust environmental perception, especially under high-speed conditions where precise synchronization is critical.

Although limited in number, highway CP datasets are gradually expanding. TUMTraf A9~\cite{zimmer2023tumtraf} captures multi-weather data from Germany’s A9 highway for 3D object detection. OTVIC~\cite{zhu2024otvic} introduces the first online-transmission, multi-view, multimodal dataset for V2I cooperative detection at 70–110 km/h. V2V4Real~\cite{xu2023v2v4real} includes 410 km of driving data, with 80\% collected in highway environments. RoScenes~\cite{zhu2024roscenes} offers 14 ring-road scenarios with up to 800-meter perception range, enabling blind-spot-free coverage. H-V2X~\cite{liu2024h} combines radar and cameras for robust BEV perception in adverse weather, marking the first real-world, large-scale highway dataset of its kind. Other datasets such as DAIR-V2X~\cite{yu2022dair}, DOLPHINS~\cite{mao2022dolphins}, and CityFlow~\cite{tang2019cityflow} also include partial highway scenarios, contributing to this domain.

\section{Datasets by Sensor Modality} \label{sec:5}

\begin{figure}[t]
\centerline{\includegraphics[width=1\linewidth]{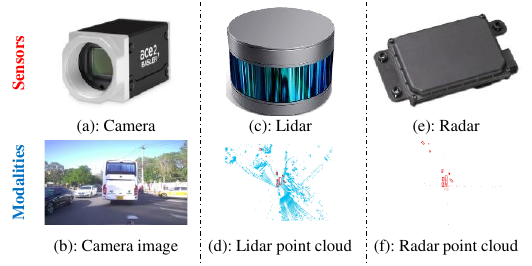}}
\caption{Representative sensor types and corresponding perception modalities. (a) Camera and (b) RGB image; (c) LiDAR and (d) LiDAR point cloud; (e) 4D Radar and (f) sparse Radar point cloud. All data samples are derived from the V2X-Radar~\cite{yang2024v2x} dataset.}
\label{fig9}
\end{figure}

Sensor modality diversity is critical for enhancing the robustness of CP systems in complex, dynamic environments~\cite{lin2024v2vformer},~\cite{yin2023v2vformer}. As shown in Fig.~\ref{fig9}, CP typically employs cameras, LiDAR, and Radar, corresponding to RGB images, dense 3D point clouds, and sparse Radar reflections. Image-only datasets are cost-effective and semantically rich but suffer from limited depth perception and sensitivity to lighting. LiDAR-only datasets provide accurate 3D localization and perform well in low-light conditions but lack semantic information and are costly to deploy. Multimodal datasets, particularly those integrating cameras and LiDAR, are prevalent due to their complementary strengths. Radar-integrated datasets have gained traction for their robustness in adverse weather~\cite{yang2024v2x}. Table~\ref{tab:sensor_comparison} summarizes the trade-offs among these modalities in terms of resolution, cost, weather resilience, and perception reliability. This section categorizes CP datasets based on sensor modality into image-only, LiDAR-only, and multimodal types.

\begin{table*}[htbp]
\centering
\caption{The Main Advantages and Disadvantages of Cameras, LiDAR, and Radar}
\begin{tabular}{>{\centering\arraybackslash}m{1.5cm} p{7cm} p{7cm}}
\toprule
\textbf{Sensor} & \textbf{Advantages} & \textbf{Disadvantages} \\
\midrule

\multirow{4}{*}{\textbf{Camera}} & 
• High resolution, capable of detecting color and texture information, good object classification ability. \newline
• Low cost, mature technology. \newline
• Recognizes traffic signs and lane markings. 
& 
• Relies on lighting conditions, reduced performance at night. \newline
• Poor distance and speed measurement capability. \newline
• Affected by weather conditions, performance degrades in adverse weather. \\

\midrule

\multirow{4}{*}{\textbf{LiDAR}} & 
• High-precision distance measurement, enables 3D environmental modeling. \newline
• Not affected by lighting conditions, performs well at night. \newline
• Wide field of view detection. 
& 
• Typically expensive. \newline
• Performance degrades in rain, snow, and affected by smoke, dust, and weather conditions. \newline
• Large data volume, complex processing. \\

\midrule

\multirow{6}{*}{\textbf{Radar}} & 
• Unaffected by weather and nighttime conditions, operates in all-weather environments. \newline
• Accurate speed measurement, long detection range. \newline
• Moderate cost, lower than LiDAR. \newline
• Strong penetration (fog, rain, dust), strong anti-interference capability. 
& 
• Low resolution. \newline
• Less sensitive to static objects, prone to false detections on metallic objects. \newline
• Sparse detection points, weak reflection from pedestrians, making detection difficult. \newline
• Cannot capture color or image information. \\

\bottomrule
\end{tabular}
    \label{tab:sensor_comparison}
\end{table*}

\subsection{Image-only Datasets}
Image-only datasets offer high-resolution visual data, supporting tasks such as object detection and semantic segmentation. However, the absence of depth information limits their performance in occlusion handling, distance estimation, and robustness under varying lighting. CityFlow~\cite{tang2019cityflow} spans large-scale urban scenarios, supporting cross-camera tracking and vehicle ReID, with extensive spatial and camera coverage. WIBAM~\cite{howe2021weakly}, captured from infrastructure-mounted cameras, focuses on 2D/3D object detection. RoScenes~\cite{zhu2024roscenes}, the largest multi-view roadside dataset to date, objects vision-based BEV perception, addressing challenges in multi-camera fusion and dense traffic scenes. Despite their effectiveness in visual tasks, image-only datasets have limited utility in high-precision localization and CP.

\subsection{LiDAR-only Datasets}
LiDAR-only datasets provide accurate spatial geometry and are widely used for 3D detection, tracking, and map construction. While superior in depth perception, the absence of semantic cues limits their effectiveness in classification. V2V-Sim~\cite{wang2020v2vnet} and CODD~\cite{arnold2021fast} offer synthetic LiDAR data for perception and registration tasks. LUCOOP~\cite{axmann2023lucoop}, a real-world dataset, supports multi-agent SLAM and urban navigation. Mixed Signals~\cite{luo2025mixed} enhances roadside cooperation through diverse LiDAR sensors. InScope~\cite{zhang2024inscope} improves spatial awareness via distributed LiDAR deployments. Although point cloud datasets excel in 3D localization, their semantic limitations drive the shift toward multimodal fusion datasets in CP research.

\subsection{Multi-modal Datasets} 

Multi-modal datasets, particularly those fusing LiDAR and cameras, are central to CP research. By combining LiDAR’s precise 3D spatial perception with camera-derived semantic richness, these datasets improve environmental understanding and robustness~\cite{hu2022data},~\cite{hu2023fault}. Recently, the integration of radar has further enhanced perception under adverse weather and for long-range detection. This section summarizes key LiDAR-Camera and LiDAR-Radar-Camera datasets.

LiDAR-Camera datasets are foundational to CP research, covering both infrastructure-based and collaborative scenarios. Ko-PER~\cite{strigel2014ko}, IPS300+~\cite{wang2022ips300+}, Rope3D~\cite{ye2022rope3d}, and A9-Dataset~\cite{cress2022a9} use roadside sensors for accurate 3D annotation, while H-V2X~\cite{liu2024h}, RCooper~\cite{hao2024rcooper}, and TUMTraf-I~\cite{zimmer2023tumtraf} address challenging conditions such as rain and nighttime. INTERACTION~\cite{zhan2019interaction} supports behavioral analysis from UAV perspectives.

Collaborative datasets like OPV2V~\cite{xu2022opv2v}, IRV2V~\cite{wei2024asynchrony}, and V2V4Real~\cite{xu2023v2v4real} enable multi-agent perception. OPV2V-N~\cite{wang2024rcdn} improves image robustness, and V2V-QA~\cite{chiu2025v2v} introduces LLM-based reasoning. In V2X contexts, DAIR-V2X-C~\cite{yu2022dair}, V2X-Seq~\cite{yu2023v2x}, and HoloVIC~\cite{ma2024holovic} support detection and planning. Simulated datasets such as V2X-Sim 2.0~\cite{li2022v2x}, V2XSet~\cite{xu2022v2x}, DOLPHINS~\cite{mao2022dolphins}, DeepAccident~\cite{wang2024deepaccident}, V2X-Real~\cite{xiang2025v2x}, and Multi-V2X~\cite{li2024multi} address a wide range of cooperative tasks.

LiDAR-Radar-Camera datasets, including OPV2V+~\cite{hu2023collaboration}, OPV2V-H~\cite{lu2024extensible}, and V2X-Radar~\cite{yang2024v2x}, enhance robustness under adverse conditions. Despite progress, challenges remain in aligning heterogeneous data and ensuring scalability across modalities—future work should emphasize cross-domain generalization and efficient multimodal fusion strategies.

\section{Perception Tasks and Metrics} \label{sec:6}

With the development of CP, various datasets have been introduced to support key perception tasks~\cite{song2023cooperative},~\cite{song2024spatial} . This chapter categorizes them by task type, including object detection, semantic segmentation, object tracking, and trajectory prediction. Object detection focuses on 3D localization of objects, while semantic segmentation enables pixel-level scene understanding. Object tracking and trajectory prediction involve continuous monitoring and future motion inference of dynamic agents. For each task, we present representative datasets and their corresponding evaluation metrics to guide researchers in selecting appropriate datasets and assessment criteria.

\subsection{Object Detection}
Object detection is one of the most fundamental tasks in CP, encompassing both 2D and 3D object detection. In V2X scenarios, 3D object detection is particularly critical, often relying on LiDAR point clouds, camera images, or multi-modal data to achieve accurate detection of vehicles, pedestrians, and other objects in dynamic environments.

\subsubsection{Datasets for Object Detection}

Numerous datasets have been developed to support V2X object detection across varying collaboration paradigms and perception modalities. Real-world datasets such as DAIR-V2X-C~\cite{yu2022dair}, V2X-Real~\cite{xiang2025v2x}, and V2V4Real~\cite{xu2023v2v4real} provide large-scale, multi-modal annotations for V2I and V2V scenarios. Simulated datasets like OPV2V~\cite{xu2022opv2v}, OPV2V+~\cite{hu2023collaboration}, and V2X-Sim 2.0~\cite{li2022v2x} offer controlled environments for benchmarking. HoloVIC~\cite{ma2024holovic} and RCooper~\cite{hao2024rcooper} emphasize occlusion-aware perception via infrastructure-side perception. Although Rope3D~\cite{ye2022rope3d} is not V2X-specific, it contributes monocular 3D benchmarks relevant to cooperative perception. Collectively, these datasets form a diverse foundation for V2X detection research.

\begin{figure}[t]
    \centering
    \includegraphics[width=\linewidth]{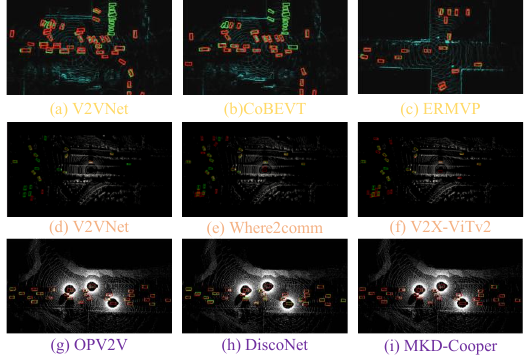}
    \caption{Qualitative detection results across three datasets: (a–c) V2V4Real with V2VNet, CoBEVT, and ERMVP~\cite{zhang2024ermvp}; (d–f) DAIR-V2X with V2VNet, Where2comm, and V2X-ViTv2~\cite{xu2024v2xv2}; (g–i) OPV2V with OPV2V, DiscoNet, and MKD-Cooper~\cite{li2023mkd}. Green boxes show ground truth; red/orange boxes indicate predictions. ERMVP and MKD-Cooper yield more accurate and robust results in complex scenes. Images reproduced from the original publications with proper citation.}
    \label{fig10}
\end{figure}

As shown in Fig.~\ref{fig10}, recent methods demonstrate improved performance across benchmarks. On V2V4Real, ERMVP~\cite{zhang2024ermvp} produces more complete and better-aligned bounding boxes than V2VNet~\cite{wang2020v2vnet} and CoBEVT~\cite{xu2022cobevt}, especially in dense scenes. On DAIR-V2X, V2X-ViTv2~\cite{xu2024v2xv2} detects more dynamic vehicles with lower spatial error than V2VNet and Where2comm~\cite{hu2022where2comm}, aided by transformer-based fusion. On OPV2V, MKD-Cooper~\cite{li2023mkd} surpasses DiscoNet~\cite{li2021learning} and V2X-ViT~\cite{xu2022v2x} in producing orientation-aware results with fewer false positives on complex road geometries. These comparisons highlight the impact of advanced fusion strategies in both real and simulated domains.

\subsubsection{Metrics for Object Detection}

3D object detection in CP is primarily evaluated using Average Precision (AP)-based metrics and task-related performance indicators~\cite{song2024robustness},~\cite{song2023graphalign++}. AP quantifies detection accuracy by computing precision at different Intersection-over-Union (IoU) thresholds \cite{gong2023feature}. Commonly used metrics include AP3D (based on 3D IoU) and APBEV (based on BEV IoU), as introduced by the KITTI benchmark~\cite{Geiger2012CVPR}.

Most datasets use AP or mAP as core metrics. RCooper~\cite{hao2024rcooper} reports AP under multiple IoUs; OPV2V~\cite{xu2022opv2v} and OPV2V+~\cite{hu2023collaboration} assess detection at 0.5 and 0.7 IoU, including within communication range. V2X-Real~\cite{xiang2025v2x} adopts lower thresholds (0.3/0.5) for scale variance. V2X-Sim 2.0~\cite{li2022v2x} emphasizes BEV-based AP. V2V4Real~\cite{xu2023v2v4real} introduces Average MegaByte (AM) to quantify communication cost under sync/async settings, while DAIR-V2X~\cite{yu2022dair} evaluates within ego-ROI under bandwidth constraints. Rope3D~\cite{ye2022rope3d} supplements AP with metrics like ACS, AOS, AAS, and AGD for comprehensive localization and shape assessment.

Beyond metrics, representative methods are benchmarked on datasets such as DAIR-V2X, V2V4Real, OPV2V, and V2XSet, as summarized in Table~\ref{tab:cp_detection_comparison}, revealing advances and ongoing challenges. In conclusion, AP and IoU remain central for evaluating CP detection, while extended metrics—e.g., BEV AP, communication overhead, and asynchronous accuracy—offer deeper realism. Future work should incorporate safety-aware evaluation to prioritize critical perception failures.

\begin{table*}[htbp]
\scriptsize
\centering
\caption{Performance Comparison of Representative Methods on Benchmark CP Datasets. All values are reported as AP@0.5 / AP@0.7. “-” indicates results not reported in the corresponding paper.}
\label{tab:cp_detection_comparison}
\renewcommand{\arraystretch}{1.15}
\begin{tabular*}{\textwidth}{@{\extracolsep{\fill}} l c c | c c c c | c c c c}
\toprule
\multirow{2}{*}{\textbf{Method}} &
\multirow{2}{*}{\textbf{Year}} &
\multirow{2}{*}{\textbf{PUB}} &
\multicolumn{4}{c|}{\textbf{Real-world Datasets}} &
\multicolumn{4}{c}{\textbf{Simulation Datasets}} \\
\cmidrule(lr){4-7} \cmidrule(lr){8-11}
& & &
\makecell{DAIR-V2X\\AP@0.5/0.7} & Ref. &
\makecell{V2V4Real\\AP@0.5/0.7} & Ref. &
\makecell{OPV2V\\AP@0.5/0.7} & Ref. &
\makecell{V2XSet\\AP@0.5/0.7} & Ref. \\
\midrule
F-Cooper~\cite{chen2019cooper} & 2019 & SEC & 0.7370/0.5600 & [96] & 0.6930/0.4320 & [96] & 0.9070/0.8220 & [97] & 0.8750/0.7390 & [97] \\
V2VNet~\cite{wang2020v2vnet} & 2020 & ECCV & 0.7097/0.4763 & [98] & 0.6470/0.3360 & [96] & 0.8970/0.7400 & [96] & 0.8710/0.6460 & [99] \\
V2VNetrobust~\cite{wang2020v2vnet} & 2021 & CoRL & 0.6610/0.4860 & [96] & 0.5500/0.3090 & [96] & 0.9420/0.8540 & [96] & - & - \\
V2X-Vit~\cite{xu2022v2x} & 2022 & ECCV & 0.7398/0.6150 & [102] & 0.6590/0.4260 & [104] & 0.9460/0.8560 & [103] & 0.9100/0.8030 & [99] \\
AttFuse~\cite{xu2022opv2v} & 2022 & ICRA & 0.7330/0.5530 & [102] & 0.7010/0.4540 & [104] & 0.9050/0.8160 & [97] & 0.8470/0.7400 & [97] \\
Where2comm~\cite{hu2022where2comm} & 2022 & NIPS & 0.7520/0.5880 & [96] & 0.7040/0.4690 & [96] & 0.9440/0.8550 & [96] & 0.9130/0.8530 & [100] \\
CoBEVT~\cite{xu2022cobevt} & 2022 & CoRL & 0.6390/0.5167 & [102] & 0.6480/0.4040 & [104] & 0.9330/0.8230 & [104] & 0.9033/0.8269 & [102] \\
DiscoNet~\cite{li2021learning} & 2021 & NIPS & 0.7370/0.5840 & [96] & \textbf{0.7360/0.4660} & [96] & 0.9160/0.7910 & [103] & 0.9078/0.8381 & [102] \\
FPV-RCNN~\cite{yuan2022keypoints} & 2022 & RAL & 0.6550/0.5050 & [103] & 0.7010/0.4790 & [96] & 0.8580/0.8400 & [103] & 0.8650/0.5630 & [99] \\
CoAlign~\cite{lu2023robust} & 2023 & ICRA & 0.7460/0.6040 & [103] & 0.7090/0.4170 & [96] & \textbf{0.9660/0.9120} & [103] & 0.9190/0.8050 & [96] \\
SCOPE~\cite{gamerdinger2024scope} & 2023 & ICCV & 0.6518/0.4989 & [105] & - & - & 0.8752/0.7505 & [105] & 0.8971/0.8062 & [105] \\
V2VFormer~\cite{lin2024v2vformer} & 2024 & TIV & - & - & 0.6540/0.3610 & [107] & 0.9170/0.8760 & [107] & - & - \\
DI-V2X~\cite{li2024di} & 2024 & AAAI & \textbf{0.7882/0.6616} & [98] & - & - & - & - & 0.9270/0.8270 & [98] \\
V2X-PC~\cite{liu2024v2x} & 2024 & arXiv & 0.7689/0.6939 & [101] & - & - & - & - & \textbf{0.9283/0.8955} & [101] \\
\bottomrule
\end{tabular*}
\end{table*}

\subsection{Semantic Segmentation}

Semantic segmentation is a crucial task in CP, aiming to achieve pixel-level classification for enhanced environmental understanding. CP semantic segmentation datasets typically integrate LiDAR point clouds and camera images to enable finer scene interpretation, thereby improving AD systems' perception in complex environments.

\subsubsection{Datasets for Semantic Segmentation}

Several datasets support semantic segmentation for CP. V2X-Sim 2.0~\cite{li2022v2x} provides multi-agent LiDAR segmentation with BEV and dynamic obstacle annotations. OPV2V~\cite{xu2022opv2v} enables BEV segmentation via LiDAR-camera fusion, while OPV2V+~\cite{hu2023collaboration} and OPV2V-N~\cite{wang2024rcdn} improve temporal alignment and image-based robustness, respectively. COMAP~\cite{yuan2021comap} offers simulated urban BEV annotations. DeepAccident~\cite{wang2024deepaccident} provides fine-grained accident scene labels. DOLPHINS~\cite{mao2022dolphins} and Adver-City~\cite{karvat2024adver} support segmentation under adverse conditions using diverse sensor setups.

Challenges remain, including inconsistent semantic labeling across agents and limited robustness under sparse or extreme conditions. Real-world datasets are also limited in scale and diversity. Future efforts should focus on large-scale, real-world, multi-agent segmentation benchmarks with strong cross-agent consistency and resilience to weather and domain shifts.

\subsubsection{Metrics for Semantic Segmentation}

Semantic segmentation in CP is typically evaluated using Pixel Accuracy (PA), mean PA (mPA), Intersection over Union (IoU), and mean IoU (mIoU)~\cite{gong2024tclanenet},~\cite{gong2024steering}. PA measures the proportion of correctly classified pixels, while mPA and mIoU address class imbalance and overall accuracy. In CP settings, evaluation extends to inter-agent consistency and modality fusion. V2X-Sim 2.0~\cite{li2022v2x} reports mIoU for semantic categories such as lanes and vehicles. OPV2V-N~\cite{wang2024rcdn} focuses on map-level segmentation using IoU. Advanced evaluation should also consider temporal stability, cross-modal consistency, and multi-agent alignment.

\subsection{Tracking and Prediction}

Object tracking and trajectory prediction focus on modeling the temporal dynamics of moving agents and are essential for AD. CP improves tracking accuracy and trajectory prediction by enabling inter-agent information sharing, particularly under occlusion and complex traffic conditions.

\subsubsection{Datasets for Tracking and Prediction}

Representative datasets for tracking and prediction span various CP paradigms. V2X-Seq~\cite{yu2023v2x} and V2XPnP-Seq~\cite{zhou2024v2xpnp} support long-sequence V2V/V2I perception with improved temporal consistency and multi-view association. DAIR-V2X-Traj~\cite{ruan2025learning} provides accurate V2I trajectory annotations, while DeepAccident~\cite{wang2024deepaccident} focuses on forecasting in safety-critical scenarios. Urban-scale datasets like CityFlow~\cite{tang2019cityflow} and INTERACTION~\cite{zhan2019interaction} enable cross-camera tracking and behavior modeling. RCooper~\cite{hao2024rcooper} captures dense roadside tracking scenes, and OPV2V~\cite{xu2022opv2v} and V2V4Real~\cite{xu2023v2v4real} benchmark V2V tracking in simulation and real-world contexts. V2X-Sim 2.0~\cite{li2022v2x} supports BEV-based multi-agent prediction. Collectively, these datasets provide diverse benchmarks for evaluating temporal perception across V2V, V2I, and infrastructure-based CP.

\subsubsection{Metrics for Tracking and Prediction}

Tracking and prediction in CP rely on metrics that evaluate detection accuracy, identity consistency, and MF. Standard CLEAR MOT metrics—MOTA, MOTP, and IDF1—remain widely used. Datasets like RCooper~\cite{hao2024rcooper} and V2V4Real~\cite{xu2023v2v4real} extend these with AMOTA and sAMOTA for scale-aware evaluation. DAIR-V2X-Traj~\cite{ruan2025learning} and V2X-Seq~\cite{yu2023v2x} report similar metrics, while V2X-Sim 2.0~\cite{li2022v2x} introduces HOTA and BPS to account for association quality and communication cost. Trajectory prediction is typically assessed using ADE and FDE, as adopted by INTERACTION~\cite{zhan2019interaction} and V2X-Seq~\cite{yu2023v2x}, with the latter also reporting Missing Rate (MR). DeepAccident~\cite{wang2024deepaccident} introduces VPQ and APA to evaluate detection-prediction synergy and accident classification. While MOTA and ADE remain core metrics, scenario-aware metrics like HOTA and VPQ offer deeper performance insights.

\section{Comprehensive Analysis of Collaborative Perception Datasets}  \label{sec:7}
 
To support robust and scalable development of CP systems, it is essential to comprehensively understand the characteristics, accessibility, and research impact of existing datasets~\cite{song2025wireless}. This section systematically analyzes all reviewed CP datasets from three perspectives: their public availability and access mechanisms, the overall quality in terms of modality, annotation, and scene diversity, and the influence of representative datasets on the research community. These insights provide valuable references for dataset selection, algorithm benchmarking, and future dataset construction.

\subsection{Availability and Access Methods of Datasets}  

Dataset accessibility plays a crucial role in advancing CP research~\cite{song2025traf}. Public availability reduces annotation costs, enhances reproducibility, and promotes innovation across the community. To evaluate current openness, we categorize CP datasets by their access methods.

Many datasets, including Ko-PER~\cite{strigel2014ko}, CityFlow~\cite{tang2019cityflow}, OPV2V~\cite{xu2022opv2v}, V2V4Real~\cite{xu2023v2v4real}, and DeepAccident~\cite{wang2024deepaccident}, support direct download from official websites without registration. Others, such as V2V-Sim~\cite{wang2020v2vnet}, V2XSet~\cite{xu2022v2x}, and CoopInf~\cite{arnold2020cooperative}, are hosted on GitHub alongside code and benchmarks, facilitating reproducible research. Datasets like DAIR-V2X-C~\cite{yu2022dair}, OTVIC~\cite{zhu2024otvic}, and WHALES~\cite{chen2024whales} require requests through cloud storage services. Some datasets, including Rope3D~\cite{ye2022rope3d}, INTERACTION~\cite{zhan2019interaction}, and RCooper~\cite{hao2024rcooper}, involve formal application processes and institutional approval. A few others, such as IRV2V~\cite{wei2024asynchrony} and OPV2V-N~\cite{wang2024rcdn}, lack clear access instructions. Detailed access types and download links are summarized in Table~\ref{tab:dataset_comparison}.

In summary, while many CP datasets are readily available, a notable portion remains behind restricted access or insufficient documentation. Enhancing openness, standardizing access procedures, and improving transparency in dataset distribution will be key to accelerating progress in CP research.

\subsection{Comprehensive Quality Analysis of Datasets} \label{sec:7.2}

\begin{figure*}[t]
\centerline{\includegraphics[width=1\textwidth]{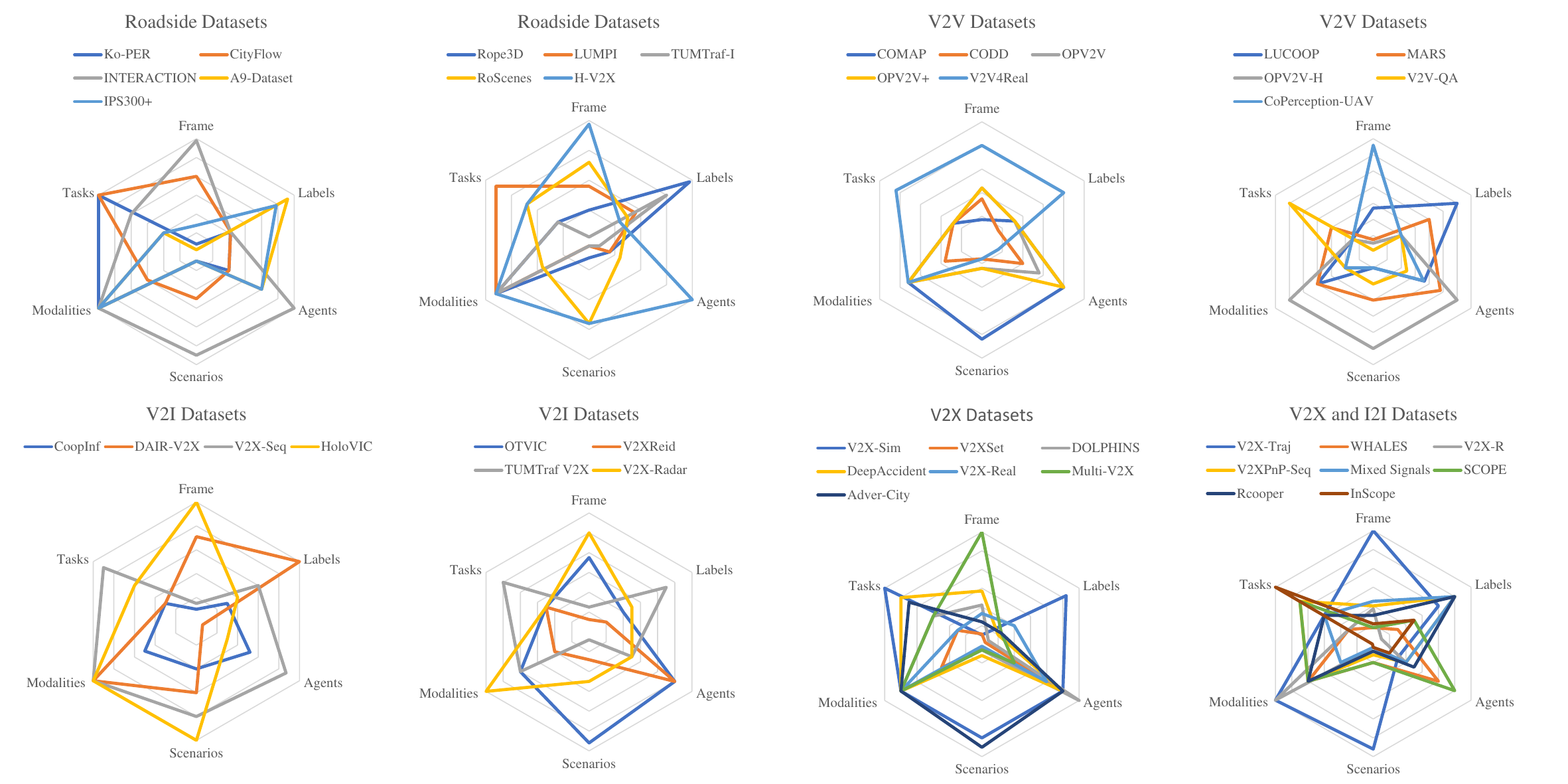}}
\caption{Qualitative radar chart analysis of CP datasets. Each subplot represents a category of datasets and compares representative datasets across six key dimensions: \textcolor{black}{Frame} (number of frames captured in each dataset), \textcolor{black}{Labels} (annotation quality and category diversity), \textcolor{black}{Agents} (number and heterogeneity of collaborating entities), \textcolor{black}{Scenarios} (scene types and environmental coverage), \textcolor{black}{Modalities} (sensor types: camera, LiDAR, radar), and \textcolor{black}{Tasks} (supported perception tasks: object detection, tracking, segmentation, prediction). The analysis is qualitative and intended to provide an intuitive comparison of dataset characteristics, highlighting their strengths and focus areas in different CP contexts.}
\label{fig11}
\end{figure*}

The quality of CP datasets directly impacts their applicability in real-world scenarios. As summarized in Table~\ref{tab:dataset_comparison} and illustrated qualitatively in Fig.~\ref{fig11}, we conduct a multi-dimensional analysis across six key aspects: data frame volume, annotation richness, agent diversity, scenario complexity, sensor modalities, and supported tasks. These dimensions reflect both dataset scalability and task coverage, providing a unified view for comparative assessment.

In terms of data volume, large-scale datasets such as RoScenes~\cite{zhu2024roscenes}, INTERACTION~\cite{zhan2019interaction}, and DAIR-V2X-Traj~\cite{ruan2025learning} contain millions of frames or annotations, supporting long-sequence modeling and prediction tasks. For sensor diversity, OPV2V~\cite{xu2022opv2v}, V2X-R~\cite{huang2024v2x}, and V2X-Radar~\cite{yang2024v2x} integrate camera, LiDAR, and radar modalities, enabling robust multimodal fusion. Regarding task support, datasets like V2X-Sim~\cite{li2022v2x}, DeepAccident~\cite{wang2024deepaccident}, and SCOPE~\cite{gamerdinger2024scope} offer multiple annotations for detection, tracking, segmentation, and prediction, enhancing their versatility for perception research. The radar charts in Fig.~\ref{fig11} further visualize dataset strengths and limitations across these dimensions within categories such as V2V, V2I, and roadside perception.

High-quality CP datasets exhibit complementary strengths: real-world datasets such as DAIR-V2X~\cite{yu2022dair} and V2X-Real~\cite{xiang2025v2x} prioritize realism and generalization, while simulation datasets like OPV2V+~\cite{hu2023collaboration} and Multi-V2X~\cite{li2024multi} offer scalability and control. The trend toward multimodal data, rich annotations, and task diversity suggests a shift to more holistic, multi-agent CP benchmarks. Standardized evaluation protocols are needed to enhance reproducibility and cross-domain comparability.

\subsection{Summary of High-Influence Datasets}
In the field of CP, several milestone datasets have gained prominence. This section summarizes key high-impact datasets across infrastructure-based, V2I, V2V, and V2X CP domains.

High-impact datasets are characterized by high citation rates and widespread benchmark adoption. In infrastructure-based perception, CityFlow~\cite{tang2019cityflow}, INTERACTION~\cite{zhan2019interaction}, and Rope3D~\cite{ye2022rope3d} are widely used for multi-camera tracking and urban understanding, supporting methods like BEVHeight~\cite{yang2023bevheight}, BEVHeight++~\cite{yang2025bevheight++}, and ViT-CoMer~\cite{xia2024vit}. In V2V, OPV2V~\cite{xu2022opv2v} enables numerous algorithms including Where2comm~\cite{hu2022where2comm}, CoBEVT~\cite{xu2022cobevt}, and OPV2V-H~\cite{lu2024extensible}. DAIR-V2X~\cite{yu2022dair} and its variants~\cite{yu2023v2x, ruan2025learning} support V2I tasks like detection and tracking, used in CoCa3D~\cite{hu2023collaboration}, HM-ViT~\cite{xiang2023hm}, and BEV-V2X~\cite{chang2023bev}. For V2X, V2X-Sim~\cite{li2022v2x} facilitates BEV detection and segmentation, widely adopted by VoxFormer~\cite{li2023voxformer} and V2X-ViT~\cite{xu2022v2x}.

In summary, datasets like CityFlow, INTERACTION, Rope3D, OPV2V, DAIR-V2X, and V2X-Sim have become cornerstones of CP research due to their broad adoption and benchmark availability. Newer datasets, including V2V4Real~\cite{xu2023v2v4real}, Multi-V2X~\cite{li2024multi}, TUMTraf V2X~\cite{zimmer2024tumtraf}, and RCooper~\cite{hao2024rcooper}, are also gaining momentum and are expected to play increasingly important roles.

\section{Challenges and Frontiers} \label{sec:8}
As research on CP continues to advance, datasets play a crucial role in facilitating multi-agent perception, V2X communication, and intelligent transportation systems. However, existing CP datasets still face numerous challenges, including insufficient data scale and diversity, limited availability, lack of standardization, domain adaptation issues, and concerns over privacy and security. Furthermore, with the rapid development of LLMs, their application in AD and CP datasets introduces new research trends. This chapter discusses these core challenges and explores potential future directions.

\subsection{Dataset Scale and Diversity}

\subsubsection{Large-Scale Datasets}
Large-scale datasets are essential for training deep learning-driven CP systems and improving their generalization ability. However, existing CP datasets are still significantly smaller than large-scale datasets in single-vehicle perception, such as the Waymo Open Dataset and nuScenes. While datasets like OPV2V, DAIR-V2X, and V2X-Sim provide relatively large sample sizes, they remain insufficient to comprehensively cover the complex and dynamic real-world traffic environment. Additionally, constructing large-scale datasets is costly and involves complex annotation processes, limiting their scalability. Future research may leverage automated annotation, synthetic data generation, and data augmentation techniques to reduce construction costs while enhancing dataset diversity and representativeness to better meet the needs of CP tasks.

\subsubsection{Datasets for Adverse Environment}
Most current CP datasets primarily cover scenes under clear weather and good lighting conditions, with limited data available for adverse weather scenarios such as heavy rain, snow, fog, and extreme lighting. Adverse environments significantly impact sensor perception capabilities, such as LiDAR experiencing severe signal attenuation in rain and cameras suffering performance degradation under strong or low-light conditions. The lack of such data leads to poor generalization of existing perception models in extreme environments. Future research should focus on collecting data in adverse weather conditions and developing weather simulation techniques in virtual environments to supplement missing data. Additionally, leveraging multimodal sensor fusion and robustness-enhancing algorithms can improve model stability under challenging conditions.

\subsubsection{Datasets for Complex Scenarios}
CP must handle complex scenarios such as high-density traffic, intersections, and lane changes. However, current datasets provide limited coverage of these scenarios. Object occlusion, dynamic interactions, and uncertainties in complex environments increase the difficulty of perception and prediction tasks. Future datasets should include diverse road types and driving behaviors to better simulate real-world driving conditions. Enhancing trajectory prediction and multi-agent interaction annotations will also provide a reliable benchmark for evaluating the performance of CP systems in complex scenarios.

\subsubsection{Datasets for Extreme Scenarios}
Extreme scenarios are critical for evaluating the safety of AD systems. However, existing datasets contain few extreme scenario samples, limiting model generalization in these key situations. For example, the DeepAccident dataset attempts to provide traffic accident data, but its coverage of accident types remains incomplete. Future research should explore generating extreme scenario data through simulation, combined with real-world data augmentation, to enhance model responsiveness in hazardous situations. Moreover, anomaly detection and Out-Of-Distribution analysis can help identify extreme scenarios, improving the safety and stability of CP systems.

\subsubsection{Geographic Diversity in Datasets}
Traffic rules, driving behaviors, and road structures vary significantly across different countries and regions, posing challenges for deploying CP models in diverse geographical areas. Many publicly available datasets are region-specific; for instance, DAIR-V2X is primarily based on Chinese road data. The lack of geographic diversity may lead to performance degradation when models are deployed in different countries. Future datasets should incorporate traffic data from various regions to enhance geographic diversity and improve cross-region generalization capabilities. Additionally, transfer learning and domain adaptation techniques can help models better adapt to different traffic environments.

\subsection{Dataset Availability and Accessibility}
Although multiple CP datasets are publicly available, some remain restricted, requiring special permissions or being limited to specific research collaborations. Furthermore, dataset storage and distribution methods are inconsistent, with some datasets hosted on unstable platforms, making access difficult for researchers. To promote research fairness and reproducibility, future efforts should focus on establishing more open data-sharing mechanisms, such as cloud-based platforms with unified data access interfaces. Additionally, standardizing dataset storage formats can improve portability and compatibility, allowing different research teams to efficiently utilize available data.

\subsection{Dataset Standardization and Benchmarking}
Current datasets exhibit significant differences in annotation formats, category definitions, and evaluation metrics, making direct comparisons between studies challenging. For example, some datasets adopt the KITTI format for 3D object detection annotations, while others follow the nuScenes or Waymo formats. Additionally, inconsistencies in coordinate systems, units, and category definitions hinder data interoperability. Establishing a unified dataset standard that includes annotation specifications, data formats, and evaluation metrics will enhance compatibility and reproducibility, promoting standardized development in CP research.

\subsection{Domain Adaptation and Generalization}
Domain adaptation is a core challenge for improving model generalization across datasets. Differences in sensors, environments, and traffic patterns often degrade performance when models are transferred across domains. In particular, CP models trained on simulation data struggle in real-world deployment due to noise, occlusion, and latency mismatches. Future research should explore unsupervised and self-supervised methods to bridge this Sim2Real gap. Promising directions include domain-aligned feature learning, simulation-aware augmentation, and leveraging large-scale pre-trained models. Additionally, Generative Adversarial Networks can help synthesize realistic cross-domain samples to support robust transfer.

\subsection{Privacy and Security}
CP involves multi-agent data sharing, raising concerns about privacy protection and data security. Malicious attackers could intercept vehicle sensor data or manipulate transmitted data to disrupt collaborative decision-making. Furthermore, datasets may contain sensitive information, such as vehicle identification details and pedestrian facial features, increasing privacy risks. To address these issues, techniques such as differential privacy and federated learning can be employed to protect user privacy while maintaining data usability. Additionally, ensuring secure V2X communication mechanisms to prevent cyber attacks is an important direction for future research.

\subsection{LLMs and Datasets}
Recently, LLMs have gained attention in the field of AD and intelligent transportation. LLMs can be used for automatic data annotation, synthetic data generation, and traffic scene analysis. However, their integration with V2X datasets is still in the early exploration stage. Key research directions include leveraging LLMs to enhance annotation efficiency and integrating LLMs for semantic understanding in multimodal perception tasks. Future trends may involve LLM-based data augmentation, AD knowledge graph construction, and the fusion of natural language with perception tasks to improve the intelligence level of V2X perception systems.

In summary, current CP datasets still face numerous challenges in terms of scale, availability, standardization, domain adaptation, and privacy security. Future directions include expanding dataset diversity and scale, establishing more comprehensive data-sharing mechanisms, promoting standardization efforts, and exploring the application of emerging technologies such as LLMs in dataset development. Advancements in these areas will further drive CP research and enhance the safety and reliability of AD systems.

\section{Conclusion} \label{sec:9}
This review provides a comprehensive overview of CP datasets for AD, categorized by collaboration paradigms, sensor modalities, data sources, environments, and supported tasks. We examine real and simulated datasets across V2V, V2I, V2X, and I2I settings, assessing their annotation quality, accessibility, scenario diversity, and benchmarking readiness. Despite notable progress, challenges remain in data scale, rare condition coverage, agent diversity, and regional generalization. Future datasets should address long-tail scenarios, broaden geographic diversity, support heterogeneous agents, and capture realistic communication constraints. Standardized annotations and unified evaluation protocols are critical for fair comparison and cross-dataset generalization. Advances such as automated labeling, high-fidelity simulation, and LLM-assisted generative modeling offer promising pathways for scalable, semantically rich datasets. We hope this review informs future dataset design and accelerates CP research toward robust real-world AD applications.

\bibliographystyle{ieeetr}    
\small\bibliography{main}

\end{document}